# A Generalized LLM-Augmented BIM Framework: Application to a Speech-to-BIM system

Ghang Lee, glee@yonsei.ac.kr
*Department of Architecture and Architectural Engineering, Yonsei University, South Korea*

Suhyung Jang, rgb000@yonsei.ac.kr
*Department of Architecture and Architectural Engineering, Yonsei University, South Korea*

Seokho Hyun, tjrgh25@yonsei.ac.kr
*Department of Architecture and Architectural Engineering, Yonsei University, South Korea*

**Abstract**

Performing building information modeling (BIM) tasks is a complex process that imposes a steep learning curve and a heavy cognitive load due to the necessity of remembering sequences of numerous commands. With the rapid advancement of large language models (LLMs), it is foreseeable that BIM tasks—including querying and managing BIM data, 4D and 5D BIM, design compliance checking, or authoring a design, using written or spoken natural language (i.e., text-to-BIM or speech-to-BIM)—will soon supplant traditional graphical user interfaces. This paper proposes a generalized LLM-augmented BIM framework to expedite the development of LLM-enhanced BIM applications by providing a step-by-step development process. The proposed framework consists of six steps: interpret-fill-match-structure-execute-check. The paper demonstrates the applicability of the proposed framework through implementing a speech-to-BIM application—NADIA-S (Natural-language-based Architectural Detailing through Interaction with Artificial Intelligence via Speech)—using exterior wall detailing as an example.

## 1 Introduction

As large language models (LLMs) rapidly evolve into large multimodal models (LMMs), the integration of these technologies into building information modeling (BIM) tasks to enhance work performance is significantly increasing. The use of generative artificial intelligence (AI) during the conceptual design phase is particularly becoming a norm in industry and academia. A recent survey by the Royal Institute of British Architects (RIBA) reported that 68% of the responding architects are already using generative AI, such as text-to-image models, for early design visualization

While the application of LLMs in BIM tasks beyond the early design phase is still in an early stage, it is foreseeable that BIM systems with natural language interfaces supported by LLMs will supplant BIM tools with traditional user interfaces in the near future. In this paper, we use the term "LLM-augmented BIM" as a general expression to indicate a task or a process of querying, generating, and managing BIM data and/or models via speech or text in natural language. We refer to the former as "speech-to-BIM" and the latter as "text-to-BIM" tasks. Existing examples of "speech-to-BIM" applications include *BIMlogiq Copilot* (BIMLOGIQ n.d.; Corke 2023) and *BIMIL AI Helper* (BIMPeers 2024). Examples of text-to-BIM applications include BIMS-GPT (Zheng and Fischer 2023), AI ChatBot of KBIM Assess Lite (Cospec Innolab n.d.), and NADIA (Natural-language-based Architectural Detailing through Interaction with Artificial Intelligence) (Jang et al. 2024). Numerous other efforts are underway to further enhance BIM workflows through LLM integration.

However, integrating an LLM into BIM is a complex process that requires careful consideration. LLM-augmented BIM is not a simple command-to-command matching process. For instance, detailing tasks often involve multiple steps, such as creating a new wall type from scratch or by duplicating an existing one, then specifying parameter values when a desired detail does not exist. After selecting a desired wall type or generating one, the user can detail a wall by replacing a wall with a low level of detail with a new one with a high level of detail. If this process is executed by natural language, unlike traditional graphical user interface (GUI)-based commands, natural language commands are often incomplete, lacking some of the necessary information. Addressing these gaps while considering the design context or the architect's intent presents a significant challenge. Additionally, guiding an LLM to produce valid and accurate results is another major challenge.

To tackle these complexities and accelerate the development of LLM-augmented BIM applications, this paper proposes a generalized LLM-augmented BIM framework, by dividing the implementation steps into six: *interpret, fill, match, structure, execute, and check,* building upon the NADIA's *interpret, fill, structure, and execute* framework.

This paper is organized as follows. The next section reviews the use of AI in architectural design in more detail and discusses research gaps. The third section explains the proposed framework and the subtasks involved in it. The fourth section demonstrates how the proposed framework can be implemented through the example of NADIA-S (NADIA via Speech). The fifth section finalizes the paper by discussing contributions, limitations, and suggestions for future work.

## 2  Performing BIM Tasks Using Natural Language

The section reviews the rapidly increasing use of natural language in BIM tasks, specifically examining its application in conceptual design and other tasks.

### 2.1 Natural-Language-Driven Conceptual Design

Among many fields in architecture, engineering, and construction, the most active area that deploys generative AI is conceptual design. Examples of this include optimizing the space layout of a hospital using generative design (Aupy 2019) and exploring potential combinations of modules for modular buildings by integrating LLMs with generative design (Gaier et al. 2023). However, these generative design cases were not language driven.

With the advancements in generative adversarial networks (GAN) and LLMs, many text-to-image models, such as Stable Diffusion (OpenAI n.d.), Midjourney (Midjourney n.d.), and DALL-E (OpenAI n.d.), have become available. And it has become increasingly common for students and architects to use these text-to-image models to generate early conceptual renderings of buildings from prompts in natural language. While these AI-generated renderings are not always as accurate as mathematically calculated renderings and may sometimes lack practical feasibility, they are rapidly gaining traction in both industry and academia due to their speed and low computational demands, often providing architects with unexpected inspirations.

Based on these advancements, a plethora of commercial products for AI-assisted early design are also available, including SketchUp Diffusion (Trimble 2023), Autodesk Forma (Autodesk n.d.), Finch 3D (Finch n.d.), and ARCHITEChTURES (ARCHITEChTURES n.d.). These are just a few examples. With rapid advances in generative AI, the number of AI tools for conceptual design is expected to grow rapidly.

### 2.2  Other Natural Language-Driven BIM Tasks

While the use of AI for conceptual design is becoming widespread, its application in other tasks remains relatively uncommon. This disparity might be due to the fact that generative AI excels in creativity, which is crucial for early-stage conceptual designs. Conceptual designs can afford a broader tolerance for inaccuracies compared to detailed designs, which require a high level of precision and correctness. Examples of LLM applications beyond non-conceptual design include *BIMlogiq Copilot* (BIMLOGIQ n.d.; Corke 2023), *BIMIL AI Helper* (BIMPeers 2024), *BIMS-GPT* (Zheng and Fischer 2023), the AI ChatBot of KBIM Assess Lite (Cospec Innolab n.d.), and NADIA (Jang et al. 2024).

Both *BIMlogiq Copilot* and *BIMIL AI Helper* deploy an LLM to execute commands for custom-developed design automation tools, eliminating the need to click buttons or write code. *BIMlogiq Copilot* (BIMLOGIQ n.d.; Corke 2023) is still in its infancy, with functionality currently limited to renaming views and generating reports.

*BIMIL AI Helper* (BIMPeers 2024) can interactively create grids and dimensions for drawings from BIM model views, as well as number and correct grid axes. Recently, *BIMIL AI Helper* added a new function to execute model checking using Revit Model Checker.

The *BIMS-GPT* framework (Zheng and Fischer 2023) was developed for natural language-based information retrieval from BIM models. As its name implies, it deployed GPT to translate a natural language query into a structured query.

The AI ChatBot of KBIM Assess Lite (Cospec Innolab n.d.) includes an LLM-enabled chatbot to automatically generate a Python code for design compliance checking from design requirements given in natural language. While it does not yet universally support the translation of all design requirements, it is very effective in generating Python code from updated building codes and regulations: i.e., generation of Python code for the design compliance checking already supported by KBIM Assess Lite.

NADIA (Jang et al. 2024) generates and modifies BIM objects interactively with the user using natural language (Jang et al. 2024; Lee et al. 2023; Lee and Jang 2023). NADIA was built on the interpret-fill-structure-execute framework. First, NADIA deploys an LLM as a *design assistant* to interpret an architect's design intent into a series of commands and then as a *design consultant* to fill in missing information with the information that the LLM was trained on. In the third step, the interpreted and supplemented information is structured in a machine-processable form. Finally, it executes requested BIM tasks using the interpreted, supplemented, and structured information from the previous steps. The performance of NADIA as a design assistant yielded 83.33% and as a design consultant 98.54%. Although these performance rates are high for general tasks, unlike conceptual designs NADIA still has room for improvement because details have to be logical and satisfy engineering requirements as close to 100% as possible.

However, these developments were conducted based on random approaches rather than on a comprehensive and structured framework that future LLM-augmented BIM system development can follow. By establishing a standardized methodology, we can accelerate the development of more sophisticated, reliable, and user-friendly LLM-augmented BIM tools, ultimately bridging the gap between conceptual promise and practical implementation.

## 3 The Generalized LLM-Augmented BIM Framework

As described in the Introduction section, the generalized LLM-augmented BIM framework consists of six steps: i.e., interpret-fill-match-structure-execute-check. Previously, we proposed the NADIA framework, which includes four steps: interpret, fill, structure, and execute. The generalized LLM-augmented BIM framework adds two more steps: match and check. Figure 1 illustrates the framework.

The interpret, fill, and structure steps are generally performed by an LLM, and the match and check steps are performed by both an LLM and a BIM tool, and the perform step by a BIM tool. Here are descriptions of each step:

1) Interpret: This step identifies a BIM task and the required subtasks and information to perform the BIM task from a natural language command input.
2) Fill: Natural language commands are often incomplete. This step supplements the missing information by leveraging the knowledge that an LLM is trained on or by prompting the user to input the missing information.
3) Match: The natural language terms and the names of objects and properties in a BIM tool may not align. Consequently, *semantic matching* between the natural language terms (or input prompt) and the BIM tool terms (or target BIM objects) must be performed.

4) Structure: BIM tools can only accommodate data or instructions given in a structured machine-readable form: e.g., structured queries, data in a machine-readable format or computer code.
5) Execute: A BIM tool executes the computer code, structured queries, or a series of commands utilizing the data passed from an LLM.
6) Check: The generative capability of an LLM is both its strength and weakness. Depending on whether the resultant outcome is valid or not, the outcome can be regarded as creative or as unreliable. The "check" step evaluates the validity of a generated solution against the given requirements such as checklists, design guidelines, and codes and regulations. The validation can be performed by running a simple checker, by connecting an LLM with a database of design requirements through retrieval-augmented generation (RAG) (Lewis et al. 2021; Yu et al. 2024)] or by interfacing a BIM tool to a design compliance checking tool, such as Solibri (Solibri n.d.), Revit Model Checker, or KBIM-Assess Lite (Cospec Innolab n.d.), and receiving the design check report from the design compliance checking tool.

These steps are represented as a linear sequential process but may be conducted in parallel or different orders. Also, some of these steps may be skipped depending on the task. For example, simple tasks such as "Rotate a model 90 degrees on the X axis" is unlikely to require "Fill" or "Check" steps. The next section demonstrates an application example.

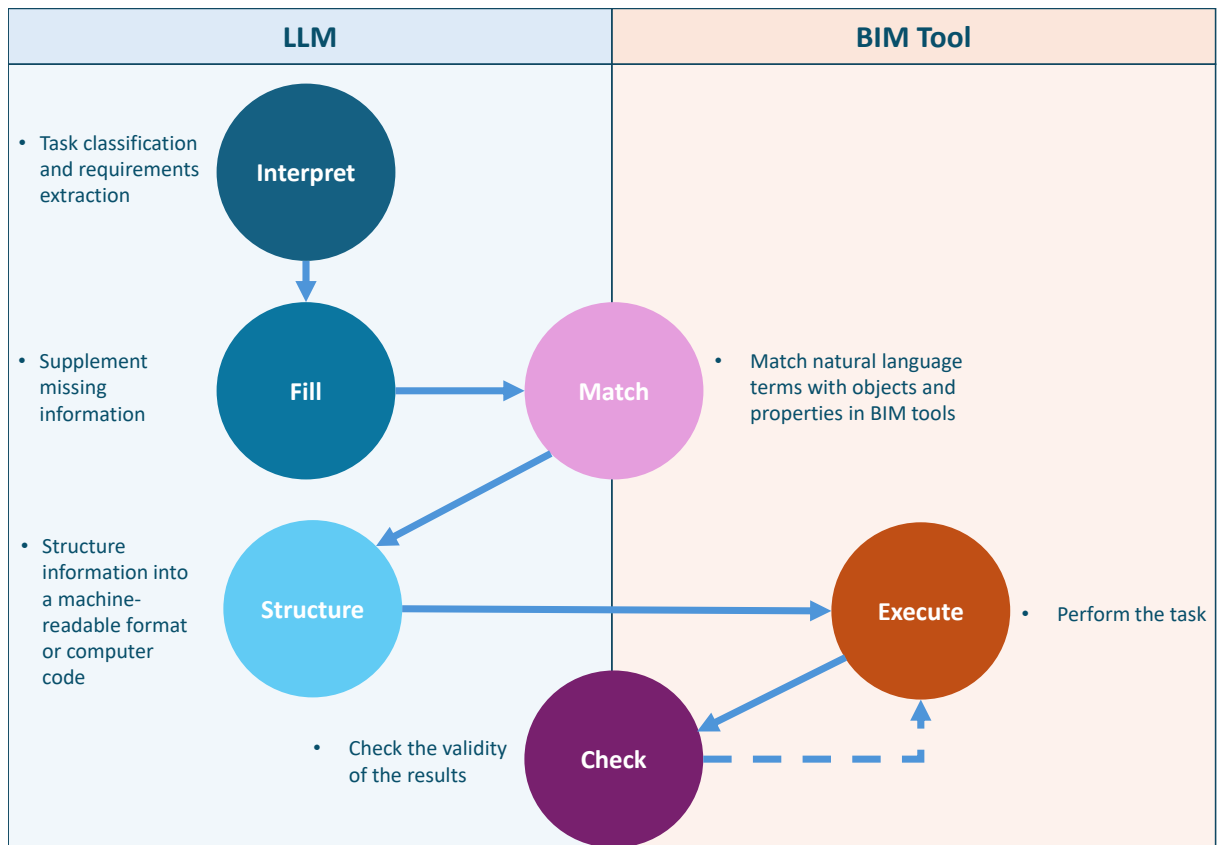

**Figure 1 The generalized LLM-augmented BIM framework**

## 4 Application example
Generating a detailed wall object and applying it to a model in current BIM tools requires more than 20 clicks and additional edits of property values. If a new version of a building information modeling (BIM) tool is released with a new menu structure or a new graphical user interface (GUI), the cognitive load required to remember all the commands and their sequence is very heavy. It is common for architects who do not use BIM tools daily to quickly lose their ability to use the tools effectively.

This section demonstrates how the proposed LLM-augmented BIM framework can be implemented through an example of NADIA-S (Natural-language-based Architectural Detailing through Interaction with Artificial Intelligence via Speech via Speech). NADIA-S is a successor to NADIA that incorporates speech recognition for interactive architectural detailing and has a much improved exception handler than its predecessor to deal more effectively with unexpected prompts.

Figure 3 shows the screenshot of the NADIA-S interface, implemented on top of Revit 2024. It is composed of the user-NAIDA dialogue history pane and the user input commands pane in addition to the conventional Revit interfaces. It uses Whisper-1 (OpenAI 2024) as a speech-to-text (STT) model. For the LLM, NADIA-S uses a combination of fine-tuned GPT-3.5-turbo-1106 and GPT-4-0613.

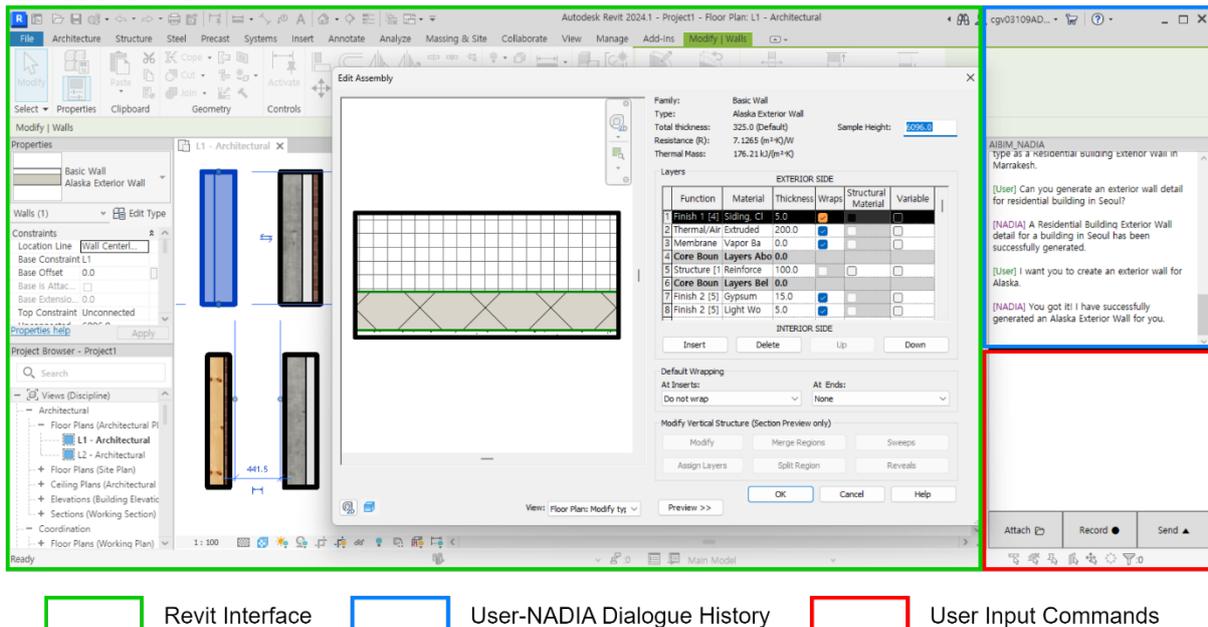

Figure 2. A screenshot of NADIA-S interface

Figure 3 illustrates how the state of data changes while going through the six steps of the proposed LLM-augmented BIM framework. As the first step, NADIA-S classifies a prompt by design task first. Examples of design tasks include "Create a new wall detail", "Place a window", "Modify a wall" and "Delete a column". Depending on the design task, an LLM is instructed to extract a set of information required to perform the task.

To address the ill-defined nature of a design problem (Eastman 1969; Simon 1973), in the second step "Fill", an LLM identifies and fills in missing information required to conduct a design task. For example, let us assume that an architect asks a system to "Create an exterior wall for Alaska." without providing details on the wall layer composition and the thickness of each layer. To generate an engineering-wise valid exterior wall detail, an LLM first needs to infer the climate conditions of Alaska and then find the typical wall details used for that weather conditions. We refer to this as the NADIA's capability as a design consultant. The automatically added information may or may not be valid. The validity of the information will be assessed in the last step, "Check."

In the third step, "Match", NADIA finds the terms in a BIM system that match the terms in the user prompt. For example, in Revit, an object library categorizes information using terms such as material, layer type, thermal conductivity, and thickness. NADIA conducts named entity recognition (NER) to find the values matching to material, layer type, thermal conductivity, and thickness from the information gained from the first and the second steps.

In the fourth step, "Structure", the wall layer composition information is translated into a machine-processable format so that the wall layer information can be passed to a BIM authoring tool to generate or update exterior wall details. In NADIA-S, the third step "Match" and the fourth step "Structure" are performed together using a single instruction prompt, such as "Return in

JSON format with 'wall_detail_name' and each layer with 'material', 'layer_type', 'thermal_conductivity' (W/m·K), and 'thickness' (mm), with exact values without units, and in order of exterior to interior layer."

In the fifth "execute" step, Revit receives the generated structured data, parses the input parameters required for generating the wall detail, and executes a function for generating the wall detail that is pre-defined within the Revit application programming interface.

As the last step "check", generated designs were validated against predefined rules. The design can be modified if it does not satisfy these rules.

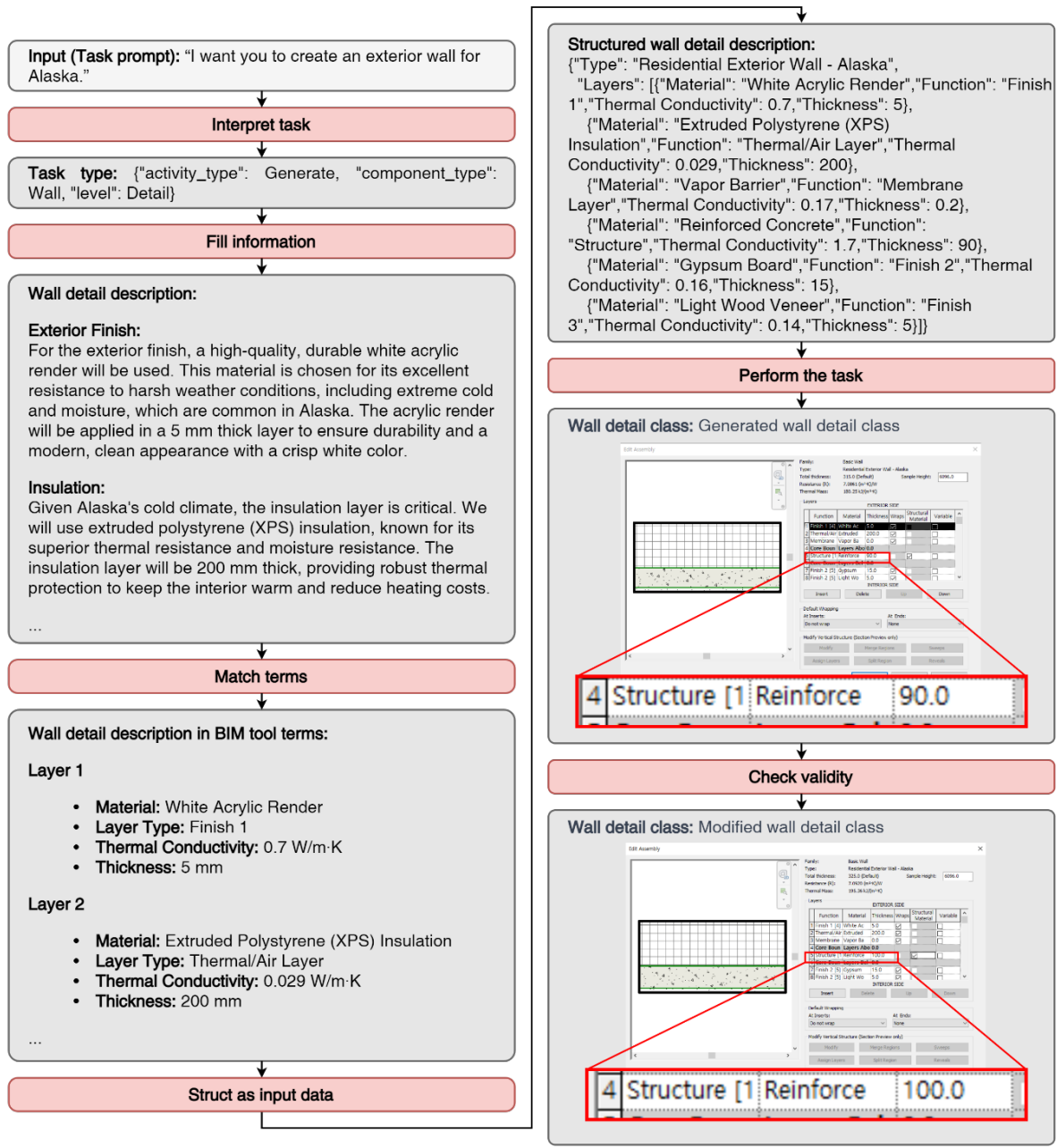

Figure 3. Six steps of BIM detail generation using NADIA-S

To measure the effectiveness of the proposed framework, we employed the same prompts used in our prior study (Jang et al. 2024). These task prompts were created by combining options for structural frame materials (reinforced concrete or timber), and wall thicknesses (140 mm or 190 mm for reinforced concrete; 140 mm or 184 mm for timber). Each unique prompt code includes abbreviations for these conditions: structural frame materials (C: reinforced concrete or T: timber), insulation methods (E: exterior or I: interior), and thicknesses (1:140 mm or 2:190 mm for reinforced concrete; 1:140 mm or 2:184 mm for timber). For example, CE1 indicates a reinforced concrete structure with exterior insulation and a minimum thickness of 140 mm. The task prompt for CE1 was, "Propose a wall detail using a reinforced concrete structure and exterior insulation method, ensuring a minimum thickness of 140 mm." Each of the eight unique task prompts was input into the previous NADIA system and executed 30 times, generating a total of 240 exterior wall details.

For the demonstration of the "Check" step, the following two requirements were applied in the experiment:

- Structural material: Evaluate if the wall detail correctly employed the requested structural material.
- Minimum structural thickness: Evaluate if the thickness of the exterior wall is over 100 mm in the case of reinforced concrete walls and 140 mm to 190 mm in the case of timber walls, applying the generally accepted minimum thickness of load-bearing walls.

The "Execute" step was repeated until automatically generated wall details satisfied all these requirements. Unsurprisingly, the performance of the proposed LLM-augmented BIM framework outperformed the previous NADIA's interpret-fill-structure-check framework (Jang et al. 2024). For structural frames, the NADIA-S developed based on the proposed framework achieved an accuracy of 100.00%, higher than the 92.50% accuracy achieved by NADIA's approach. In terms of minimum structural thickness, the proposed framework achieved an accuracy of 100.00%, surpassing NADIA's accuracy of 91.66%.

In summary, although the experiment was conducted using a rudimentary example, it clearly demonstrated the potential of the generalized LLM-augmented BIM framework. To apply more complex rules, we are experimenting with retrieval-augmented generation.

## 5 Conclusion

As LLM technologies, along with AI, continue to evolve, their role in AEC will likely expand, reducing the cognitive load on professionals to learn and remember new or frequently updated tools and sequences of commands and, eventually, enabling more efficient and precise execution of BIM tasks. However, a generalized and structured approach for developing a system that integrates LLMs and BIM tools to perform BIM tasks in natural language has not yet been explored. This study proposes a generalized LLM-augmented BIM framework, which divides the implementation steps into interpret-fill-match-structure-execute-check. It builds upon NADIA's interpret-fill-structure-execute framework.

This study demonstrated the potential of the generalized LLM and BIM integration framework through the implementation of NADIA-S. While the experiment was simple, it clearly demonstrated that the proposed framework could enhance the future speech-to-BIM application development. Future studies on the following four areas can benefit the development of mature LLM-augmented BIM applications.

- Semantic Matching: While essential, semantic matching can be complex, especially in the context of domain-specific language used in BIM. Exploring advanced natural language processing techniques could enhance accuracy.
- LLM Capabilities: The effectiveness of the framework heavily relies on the capabilities of the LLM used. Research into LLMs specifically trained on BIM data or architecture-related tasks could improve performance, particularly for the first and the second steps, "Interpret" and "Fill".
- Validity checking: There have been quite a few studies on BIM data quality checking and design compliance checking. However, evaluating the outcomes of an AI is challenging and remains much to be explored.
- User Interaction: The framework could benefit from developing and incorporating advanced user feedback mechanisms to refine the generated BIM models and improve overall user experience.

Although the effectiveness of the proposed framework was demonstrated with a focus on wall detailing via speech-to-BIM, it can also be applied to any natural-language-driven BIM tasks, such as data querying, design compliance checking, data management, data analysis, question answering, cost and schedule estimation, and more. We hope that the proposed framework will accelerate the development of LLM-augmented BIM systems, ultimately enhancing the usability and efficiency of BIM tools for professionals.

## Acknowledgements
This work is supported in 2024 by the Korea Agency for Infrastructure Technology Advancement (KAIA) grant funded by the Ministry of Land, Infrastructure and Transport (Grant RS-2021-KA163269).